\documentclass[11pt,a4paper]{article}
\usepackage{authblk}
\usepackage[hyperref]{naaclhlt2018}
\usepackage{times}
\usepackage{latexsym}
\usepackage{graphicx}
\usepackage{url}
\usepackage{amsthm}
\usepackage{bbding}
\usepackage{pifont}
\usepackage{wasysym}
\usepackage{amssymb}

\aclfinalcopy 

\title{Suggestion Mining from Online Reviews using ULMFiT}

\author{
Sarthak Anand,\textsuperscript{\rm 3}
Debanjan Mahata,\textsuperscript{\rm 1}
Kartik Aggarwal, \textsuperscript{\rm 3}
Laiba Mehnaz,\textsuperscript{\rm 2}
Simra Shahid,\textsuperscript{\rm 2} \\
Haimin Zhang,\textsuperscript{\rm 1}
Yaman Kumar,\textsuperscript{\rm 5}
Rajiv Ratn Shah,\textsuperscript{\rm 4}
Karan Uppal\textsuperscript{\rm 1} \\ \\
\textsuperscript{\rm 1}Bloomberg, USA,
\textsuperscript{\rm 2}DTU-Delhi, India,
\textsuperscript{\rm 3}NSIT-Delhi, India,
\textsuperscript{\rm 4}IIIT-Delhi, India,
\textsuperscript{\rm 5}Adobe, India,\\
sarthaka.ic@nsit.net.in, dmahata@bloomberg.net, kartik.mp.16@nsit.net.in, \\ laibamehnaz@dtu.ac.in, simrashahid\_bt2k16@dtu.ac.in, hzhang449@bloomberg.net, \\ykumar@adobe.com, rajivratn@iiitd.ac.in, kuppal8@bloomberg.net
}

\date{}

\begin{document}
\maketitle
\begin{abstract}
In this paper we present our approach and the system description for Sub Task A of SemEval 2019 Task 9: Suggestion Mining from Online Reviews and Forums. Given a sentence, the task asks to predict whether the sentence consists of a suggestion or not. Our model is based on Universal Language Model Fine-tuning for Text Classification. We apply various pre-processing techniques before training the language and the classification model. We further provide detailed analysis of the results obtained using the trained model. Our team ranked 10th out of 34 participants, achieving an F1 score of 0.7011. We publicly share our implementation\footnote{\url{https://github.com/isarth/SemEval9_MIDAS}}.
\end{abstract}

\section{Introduction and Background}

Suggestion mining can be defined as the process of identifying and extracting sentences from unstructured text that contain suggestion \cite{negi2018open}. Suggestions in the form of unstructured text could be found in various social media platforms, discussion forums, review websites and blogs. They are often expressed in the form of advice, tips, recommendations, warnings, things to do, and various other forms in an explicit as well as an implicit way.

Identifying and retrieving suggestions from text can be useful in an industrial setting for enhancing a product, summarizing opinions of the consumers, giving recommendations and as an aid in decision making process \cite{jijkoun2010mining}. For normal users of online platforms it could help in seeking advice related to general topics of interest like travel, health, food, shopping, education, and many more. Given the abundance of textual information in the Internet about a variety of topics, suggestion mining is certainly an useful task interesting to researchers working in academia as well as industry. 

Most of the previous efforts in the direction of understanding online opinions and reactions have been limited to developing methods for areas like sentiment analysis and opinion mining \cite{medhat2014sentiment, baghel2018kiki, kapoor2018mind, mahata2018detecting, mahata2018phramacovigilance, jangid2018aspect, meghawat2018multimodal, shah2017multimodal}. Mining and understanding suggestions can open new areas to study consumer behavior and tapping nuggets of information that could be directly linked with the development and enhancement of products \cite{brun2013suggestion, dong2013automated, ramanand2010wishful}, improve customer experiences \cite{negi2015towards}, and aid in understanding the linguistic nuances of giving advice \cite{wicaksono2013automatic}. 

Suggestion mining is a relatively new domain and is challenged by problems such as \textit{ambiguity in task formulation and manual annotation}, \textit{understanding sentence level semantics}, \textit{figurative expressions}, \textit{handling long and complex sentences}, \textit{context dependency}, and \textit{highly imbalanced class distribution}, as already mentioned by \cite{negi2018open}. Similar problems are also observed in the dataset shared by the organizers for the SemEval task, as it is obtained from a real-world application comprising of suggestions embedded in unstructured textual content.

\noindent \textit{Problem Definition} - The problem of suggestion mining as presented in the SemEval 2019 Task 9 \cite{negi2019semeval}, is posed as a binary classification problem and could be formally stated as:

\noindent \textit{Given a labeled dataset $D$ of sentences, the objective of the task is to learn a classification/prediction function that can predict a label $l$ for a sentence $s$, where $l \in \{suggestion, non suggestion\}$.}

\noindent \textit{Our Contributions} - Some of the contributions that we make by participating in this task are:

\noindent $\bullet$ To our knowledge we are the first one to use Universal Language Model Fine-tuning for Text Classification (ULMFiT) \cite{DBLP:journals/corr/abs-1801-06146}, for the task of suggestion mining and show the effectiveness of transfer learning. 

\noindent $\bullet$ We perform an error analysis of the provided dataset for Sub Task A, as well as the predictions made by our trained model.

Next, we give a detailed description of our system and the experiments performed by us along with explaining our results.

\section{Experiments}

\subsection{Dataset\label{data}}
The dataset used in all our experiments was provided by the organizers of the task and consists of sentences from a suggestion forum annotated by humans to be a \textit{suggestion} or a \textit{non-suggestion}. Suggestion forums are dedicated forums used for providing suggestions on a specific product, service, process or an entity of interest. The provided dataset is collected from uservoice.com\footnote{https://www.uservoice.com/}, and consists of feedback posts on Universal Windows Platform.
Only those sentences are present in the dataset that explicitly expresses suggestions, for example - \textit{Do try the cupcakes from the bakery next door}, instead of those that contain implicit suggestions such as - \textit{I loved the cup cakes from the bakery next door} \cite{negi2018open}.

\begin{table}[htbp]
\begin{center}
\begin{tabular}{|l|r|l|}
\hline  \bf Label & \bf Train & \bf Trial \\ \hline
\bf Suggestion &  2085 &  296 \\
\bf Non Suggestion &  6415 & 296 \\
\hline
\end{tabular}
\end{center}
\caption{\label{train-data} Dataset Distribution for Sub Task A - Task 9:  Suggestion Mining from Online Reviews.}
\label{data-distribution}
\end{table}

For Sub Task A, the organizers shared a training and a validation dataset whose label distribution (\textit{suggestion} or a \textit{non-suggestion}) is presented in Table \ref{data-distribution}. The unlabeled test data on which the performance of our model was evaluated was also from the same domain. As evident from Table \ref{data-distribution}, there is a significant imbalance in the distribution of training instances that are \textit{suggestions} and \textit{non-suggestions}, which mimics the distributions of these classes in the real-world datasets. Although the dataset was collected from a suggestion forum and is expected to have a high occurrence of suggestions, yet the imbalance is more prominent due to the avoidance of implicit suggestions.

\subsection{Dataset Preparation}

\begin{table*}[htbp]
\centering
\begin{tabular}{|l|l|}
\hline
\multicolumn{1}{|c|}{\textbf{\begin{tabular}[c]{@{}c@{}}Text Snippet\\ before Pre-processing\end{tabular}}} & \multicolumn{1}{c|}{\textbf{\begin{tabular}[c]{@{}c@{}}Text Snippet \\ after Pre-processing\end{tabular}}} \\ \hline
ie9mobile does not do this :( & ie mobile does not do this \textless{}emsad\textgreater{} \\ \hline
\begin{tabular}[c]{@{}l@{}}For example if you want a feed for every Tumblr \\ feed containing the hashtags `` ``\#retail \#design " "; \\ `` ``http://www.tumblr .com/tagged/retail+ design""; \\ would be a feedly feed."\end{tabular} & \begin{tabular}[c]{@{}l@{}}For example if you want a feed \\ for every tumblr feed containing \\ the hashtags \textless{}hashtag\textgreater retail \\ \textless{}hashtag\textgreater design \textless{}url\textgreater would \\ be a feedly feed\end{tabular} \\ \hline
\end{tabular}
\caption{Text snippet from the dataset before and after applying pre-processing steps.}
\label{pre-processing}
\end{table*}

Before using the provided dataset for training a prediction model, we take steps to prepare it as an input to our machine learning models. We primarily use Ekphrasis\footnote{https://github.com/cbaziotis/ekphrasis} for implementing our pre-processing steps. Some of the steps that we take are presented in this section.

\subsubsection{Tokenization}
Tokenization is a fundamental pre-processing step and could be one of the important factors influencing the performance of a machine learning model that deals with text. As online suggestion forums include wide variation in vocabulary and expressions, the tokenization process could become a challenging task. Ekphrasis ships with custom tokenizers that understands expressions found in colloquial languages often used in forums and has the ability to handle hashtags, dates, times, emoticons, besides standard tokenization of English language sentences. We also had to tokenize certain misspellings and slangs (eg. ``I'm", ``r:are") after carefully inspecting the provided dataset.

\subsubsection{Normalization}
After tokenization, a range of transformations such as word-normalization, spell correction and segmentation are applied to the extracted tokens. During word-normalization, URLs, usernames, phone numbers, date, time, currencies and  special type of tokens such as hashtags, emoticons, censored words etc. are recognized and replaced by masks (eg. $<$date$>$, $<$hashtag$>$, $<$url$>$). These steps results in a reduction in the vocabulary size without the loss of informative excerpts that has signals for expressing suggestions. This was validated manually by analyzing the text after applying the different processing steps. Table \ref{pre-processing} shows an example text snippet and its form after the application of the pre-processing steps.


\subsubsection{ Class Imbalance }
As already pointed in Section \ref{data}, \textit{class imbalance} is a prevalent challenge in this domain and is reflected in the provided dataset. We use oversampling technique in order to tackle this challenge. We duplicate the training instances labeled as \textit{suggestions} and boost their number of occurrences exactly to double the amount present in the original dataset.  

\section{Model Architecture Training and Evaluation}
We show the effectiveness of transfer learning for the task of suggestion mining by training Universal Language Model Fine-tuning for Text Classification (ULMFiT) \cite{DBLP:journals/corr/abs-1801-06146}. One of the main advantages of training ULMFiT is that it works very well for a small dataset as provided in the Sub Task A and also avoids the process of training a classification model from scratch. This avoids overfitting. We use the fast.ai\footnote{https://docs.fast.ai/text.html} implementation of this model.

The ULMFiT model has mainly two parts, the \textit{language model} and the \textit{classification model}. The language model is trained on a Wiki Text corpus to capture general features of the language in different layers. We fine tune the language model on the training, validation and the evaluation data. Also, we additionally scrap around two thousand reviews from the Universal Windows Platform for training our language model. After analysis of the performance we find optimal parameters to be:
 
\begin{itemize}
  \item BPTT: 70, bs: 48.
  \item Embedding size: 400, hidden size: 1150, num of layers: 3
\end{itemize}

We also experiment with MultinomialNB, Logistic Regression, Support Vector Machines, LSTM. For LSTM we use fasttext word embeddings\footnote{https://fasttext.cc/docs/en/pretrained-vectors.html} having 300 dimensions trained on Wikipedia corpus, for representing words.

Table \ref{performance-comparison}, shows the performances of all the models that we trained on the provided training dataset. We also obtained the test dataset from the organizers and evaluated our trained models on the same. The ULMFiT model achieved the best results  with a F1-score of 0.861 on the training dataset and a F1-score of 0.701 on the test dataset. Table \ref{performance} shows the performance of the top 5 models for Sub Task A of SemEval 2019 Task 9. Our team ranked 10th out of 34 participants.

\begin{table}[htbp]
\centering
\scalebox{0.9}{
\begin{tabular}{|c|c|c|}
\hline
\textbf{Model} & \textbf{F1 (train)} & \textbf{F1 (test)} \\ \hline
\textbf{\begin{tabular}[c]{@{}c@{}}Multinomial Naive Bayes\\ (using Count Vectorizer)\end{tabular}} & 0.641 & 0.517 \\ \hline
\textbf{\begin{tabular}[c]{@{}c@{}}Logistic Regression\\ (using Count Vectorizer)\end{tabular}} & 0.679 & 0.572 \\ \hline
\textbf{\begin{tabular}[c]{@{}c@{}}SVM (Linear Kernel)\\ (using TfIdf Vectorizer)\end{tabular}} & 0.695 & 0.576 \\ \hline
\textbf{\begin{tabular}[c]{@{}c@{}}LSTM\\ (128 LSTM Units)\end{tabular}} & 0.731 & 0.591 \\ \hline
\textbf{Provided Baseline} & 0.720 & 0.267 \\ \hline
\textbf{ULMFit*} & 0.861 & 0.701 \\ \hline
\end{tabular}}
\caption{Performance of different models on the provided train and test dataset for Sub Task A.}
\label{performance-comparison}
\end{table}

\begin{table}[htbp]
\centering
\begin{tabular}{|c|c|c|}
\hline
\textbf{Ranking} & \textbf{Team Name} & \textbf{\begin{tabular}[c]{@{}c@{}}Performance\\ (F1)\end{tabular}} \\ \hline
\textbf{1} & OleNet & 0.7812 \\ \hline
\textbf{2} & \multicolumn{1}{l|}{ThisIsCompetition} & 0.7778 \\ \hline
\textbf{3} & m\_y & 0.7761 \\ \hline
\textbf{4} & yimmon & 0.7629 \\ \hline
\textbf{5} & NTUA-ISLab & 0.7488 \\ \hline
\textbf{10} & \textbf{MIDAS (our team)} & \textbf{0.7011*} \\ \hline
\end{tabular}
\caption{Best performing models for SemEval Task 9: Sub Task A.}
\label{performance}
\end{table}

\section{Error Analysis}
\begin{figure}[h!t] 
\centering
\includegraphics[scale=0.60]{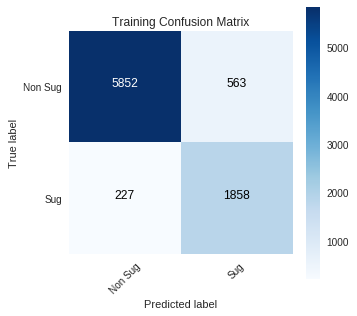}
\caption{Confusion matrix training data}
\label{fig:X}
\end{figure}

In this section, we analyse the performance of our best model (ULMFiT) on the training data as shown by the confusion matrix presented in Figure \ref{fig:X}. We specially look at the predictions made by our model that falls into the categories of False Positive and False Negative, as that gives us insights into the instances which our model could not classify correctly. We also present some of the instances that we found to be wrongly labeled in the provided dataset.

%
%
%
%
%
\noindent {\bf False Positives (Labeled or predicted wrongly as suggestion)}
Some examples that seems incorrectly labeled as suggestion in training data are given below: 
\begin{itemize}
\item{\bf Id 2602}: Current app extension only supports loading assets and scripts.
\item {\bf Id 3388}: One is TextCanvas for Display and Editing both Text and Inking.
\item {\bf Id 0-1747}: Unfortunately they only pull their feeds from google reader
\end{itemize}

\noindent Some examples that are incorrectly predicted by the model as suggestions are:
\begin{itemize}
\item {\bf Id 1575}: That's why I'm suggesting a specialized textbox for numbers.
\item {\bf Id 1462}: If you have such limits publish them in the API docs.
\item {\bf Id 1360-2}: Adding this feature will help alot.
\end{itemize}

\noindent {\bf False Negatives (Labeled or predicted wrongly as Non Suggestion)}
Some examples that seems incorrectly labeled as non suggestion in the training data:
\begin{itemize} 
\item {\bf Id 0-1594}: Please consider adding this type of feature to feedly.
\item {\bf Id 3354}: Please support the passing of all selected files as command arguments.
\item {\bf Id 0-941}: Microsoft should provide a SDK for developers to intergate such feedback system in their Apps.
\end{itemize}
Some examples that are incorrectly predicted by the model as non-suggestions:
\begin{itemize} 
\item {\bf Id 0-757}: Create your own 3d library.
\item {\bf Id 834-15}: Please try again after a few minutes" in Firefox. 
\item {\bf Id 4166}: I want my user to stay inside my app.
\end{itemize}

We also find that \textbf{77\%} of the false positives have keywords (\textit{want, please, add, support, would, could, should, need}), with \textbf{would} being highest i.e. around 30\%.

\section{Conclusion and Future Work}
In this work we showed how transfer learning could be used for the task of classifying sentences extracted from unstructured text as suggestion and non-suggestions. Towards this end we train a ULMFiT model on the dataset (only Sub Task A) provided by the organizers of the SemEval 2019 Task 9 and rank 10th in the competition out of 34 participating teams.

In the future we would like to experiment and show the effectiveness of our trained model in Sub Task B where the training dataset remains the same, but the test dataset consists of suggestions from a different domain. It would be interesting to see how our model performs in predicting out-of-domain suggestions and show the ability of the ULMFiT model to fine-tune itself to a completely new domain with the already existing pre-trained model. Another interesting area would be to explore Multi Task Learning models and see how the domain of suggestion mining could get benefited by borrowing weights from models trained on other related tasks and similar tasks across different domains.

\bibliography{semeval2018}
\bibliographystyle{acl_natbib}

\end{document}